\documentclass{article}

\date{}
\usepackage[utf8]{inputenc} 
\usepackage[T1]{fontenc}    
\usepackage{hyperref}       
\usepackage{url}            
\usepackage{booktabs}       
\usepackage{amsfonts}       
\usepackage{nicefrac}       
\usepackage{microtype}      
\usepackage{lipsum}		
\usepackage{upgreek}
\usepackage{subfig, graphicx}
\usepackage{natbib}
\usepackage{doi}
\usepackage[acronym,shortcuts]{glossaries}
\usepackage{url,hyperref,lineno,microtype}
\usepackage{cleveref}
\usepackage{threeparttable}
\usepackage{color}
\usepackage{array}
\usepackage{longtable}
\usepackage{mathtools}
\usepackage{multirow}
\usepackage{authblk}
\usepackage{rotating}

\crefname{figure}{Fig.}{Fig.}

\newacronym{MOT}{MOT}{Multiple Object Tracking}
\newacronym{MOTS}{MOTS}{Multiple Object Tracking and Segmentation}
\newacronym{CNN}{CNN}{Convolutional Neural Network}
\newacronym{DNN}{DNN}{Deep Neural Network}
\newacronym{GCN}{GCN}{Graph Convolutional Network}
\newacronym{IoU}{IoU}{Intersection over Union}
\newacronym{GIoU}{GIoU}{Generalized Intersection over Union}
\newacronym{AI}{AI}{Artificial Intelligence}
\newacronym{FPS}{FPS}{Frames Per Second}
\newacronym{MOTA}{MOTA}{Multiple Object Tracking Accuracy}
\newacronym{HOTA}{HOTA}{Higher Order Tracking Accuracy}

\newacronym{SLAM}{SLAM}{Simultaneous Localization and Mapping}
\newacronym{GNSS}{GNSS}{Global Navigation Satellite System}
\newacronym{SOTA}{SOTA}{State-Of-The-Art}
\newacronym{IDR}{IDR}{Identification Recall}
\newacronym{IDP}{IDP}{Identification Precision}
\newacronym{DetA}{DetA}{Detection Accuracy Score}
\newacronym{AssA}{AssA}{Association Accuracy Score}
\newacronym{AssRe}{AssRe}{Association Recall}
\newacronym{AssPr}{AssPr}{Association Precision}

\newacronym{FD}{FD}{Fourier Descriptor}
\newacronym{HM}{HM}{Hu Moments}

\newacronym{AR}{AR}{Average Recall}
\newacronym{AP}{AP}{Average Precision}

\newacronym{RA}{RA}{Ratio of Axis}

\title{Segmentation and Tracking of Vegetable Plants by Exploiting Vegetable Shape Feature for Precision Spray of Agricultural Robots}

\author[1]{Nan Hu}
\author[1]{Daobilige Su\thanks{Corresponding Author, Email: sudao@cau.edu.cn}}
\author[1]{Shuo Wang}
\author[1]{Xuechang Wang}
\author[1]{Huiyu Zhong}
\author[1]{Zimeng Wang}
\author[2]{Yongliang Qiao}
\author[1]{Yu Tan}
\affil[1]{College of Engineering, China Agricultural University, Beijing 100083, China}
\affil[2]{Australian Institute for Machine Learning (AIML), The University of Adelaide, 5005, Australia}

\begin{document}
\maketitle

\begin{abstract}
With the rapid growth of the world population, shortage in labor force and change in the global climate, increasing food demand and food security have become the top priorities that agriculture needs to solve urgently. The rapid development of artificial intelligence and robotics technologies make it possible as a key part of agricultural production.
With the increasing deployment of agricultural robots, the traditional manual spray of liquid fertilizer and pesticide is gradually being replaced by agricultural robots. Compared to conventional spray methods which adopt large angle spray nozzles and undifferentiated spray strategy, precision target spray has gained increasing attention as an important concept of precision agriculture, which brings in a more economical and environmentally friendly solution.
For robotic precision spray application in vegetable farms, accurate plant phenotyping through instance segmentation and robust plant tracking are of great importance and a prerequisite for the following spray action.
Regarding the robust tracking of vegetable plants, to solve the challenging problem of associating vegetables with similar color and texture in consecutive images, in this paper, a novel method of \ac{MOTS} is proposed for instance segmentation and tracking of multiple vegetable plants. 
In our approach, contour and blob features are extracted to describe unique feature of each individual vegetable, and associate the same vegetables in different images.
By assigning a unique ID for each vegetable, it ensures the robot to spray each vegetable exactly once, while traversing along the farm rows. 
Comprehensive experiments including ablation studies are conducted, which prove its superior performance over two \ac{SOTA} \ac{MOTS} methods. 
The proposed method achieves a \ac{HOTA} score higher than 70 and an \ac{AssPr} score higher than 80. The execution speed of the method reaches 29 \ac{FPS} on a consumer level hardware, which satisfies the real-time operation.
Compared to the conventional \ac{MOTS} methods, the proposed method is able to re-identify objects which have gone out of the camera field of view and re-appear again using the proposed data association strategy, which is important to ensure each vegetable be sprayed only once when the robot travels back and forth.
Although the method is tested on lettuce farm, it can be applied to other similar vegetables such as broccoli and canola. Both code and the dataset of this paper is publicly released for the benefit of the community: \textcolor{red}{\url{https://github.com/NanH5837/LettuceMOTS}}.

\end{abstract}

\section{Introduction}
\label{sec:introduction}

With the world population growth and climate change, the urgent need for food safety and sustainable production have put forward higher requirements for agriculture. With the shortage of labor force and the limited area of arable land, artificial intelligence and robotics technologies have gained significant attention in agriculture recently.
Agricultural robots are increasingly being deployed for tasks including weeding~\citep{ J_RAL2018:McCool}, crop and weed detection~\citep{J_JFR2017:Bac}, and pesticide and fertilizer application~\citep{J_JFR2017:Adamides} \textit{etc.} 
Application of liquid pesticide and fertilizer is an important process in planting vegetables. Conventional spraying techniques tend to apply liquid pesticide and fertilizer uniformly on vegetable farms, which not only leads to a waste of chemical, but also is not environmentally friendly.
In comparison, precision spray of individual plant can effectively resolve the above problem~\citep{J_IJRR2017:Chebrolu}. Images captured by the vision sensor of the robot can be used to detect vegetables and compute the location of the each vegetable, which guides the robot to apply chemical to each individual vegetable.
Accurate detection of vegetables is prerequisite for robotic precision spray. However, only detection of vegetables is usually not enough for robotic precision spray. With only detection results of vegetables, robots usually have to move with a fixed distance every time and spray all detected vegetables at each stop. In this way, robots also have to make sure that there is no overlap between two camera field of views, as well as no vegetable is missed between two consecutive camera filed of views. However, ensuring such a fixed distance is normally difficult, and a vegetable plant is likely to either get sprayed more than once, or be missed by the robot. 
Another way to handle this problem is to use \ac{GNSS} information or \ac{SLAM} methods to mark down geo-information of vegetables, so that each plant is assigned to a unique ID. However, \ac{GNSS} antenna introduces additional cost and is unstable inside greenhouse. Visual \ac{SLAM} algorithms are generally not robust in the semi-structured agricultural environment. 

\begin{figure*}[ht]
  \begin{center}
  \includegraphics[width=\columnwidth]{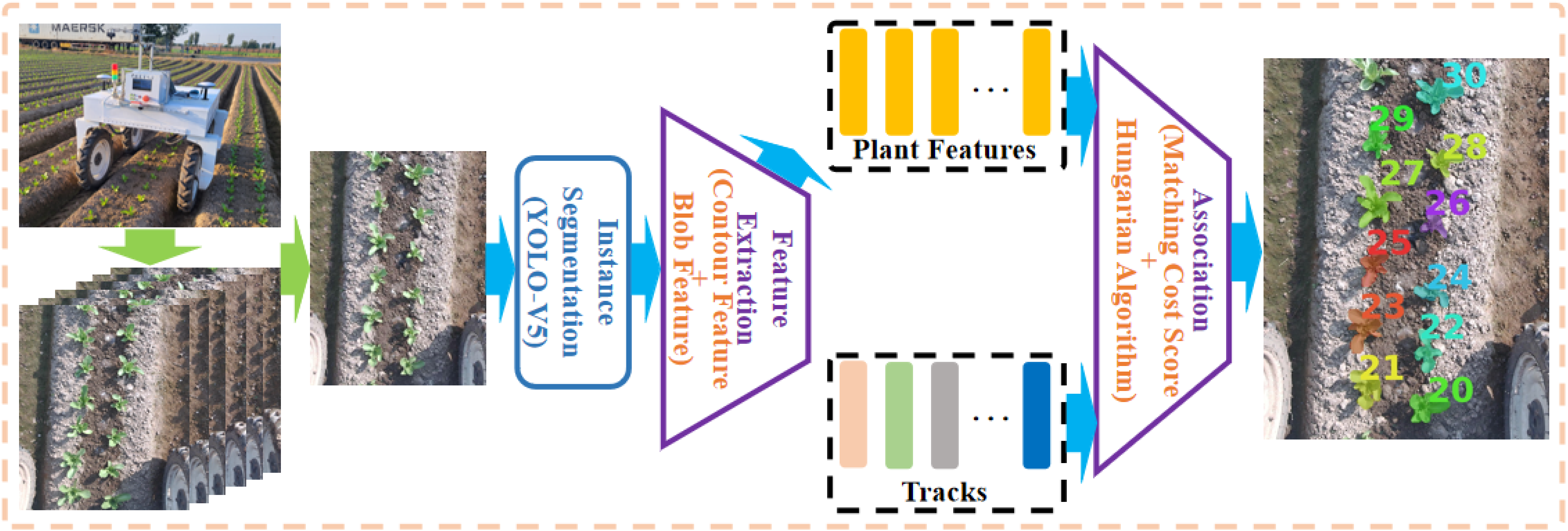}%
  \caption{Overview of the proposed method. A downward facing RGB camera is attached in front of the agricultural robot VegeBot, which collects a sequence of images while traveling through a lettuce farm. The proposed method segments and tracks each vegetable instance. Vegetable plants are firstly segmented with YOLOv5 instance segmentation net. Then, shape features consisting of contour and blob characteristics of plants are extracted to tackle the challenging data association problem of plants with similar color and texture. Based on the defined matching cost and Hungarian algorithm, accurate and robust tracking of plants is obtained.
  The details of the proposed method are described in \cref{sec:method}.}
  \label{fig:overview}
  \end{center}
\end{figure*}

To tackle this challenging problem in a better way, robots have to solve the data association problem for the detected vegetables on the consecutively captured images. When the same vegetable plants on the consecutively captured images are correctly associated to each other and identified as one,  unique IDs are assigned to them and the robot can ensure to spray each plant only once. In this way, it forms the classic \ac{MOT}\citep{C_ICIP2016:Bewley, C_ECCV2022:Zhang} or \ac{MOTS} problem \citep{C_CVPR2019:Voigtlaender, J_TIP2022:Gao}. Compared to \ac{MOT} methods\citep{J_FPS2022:Hu}, which use bounding boxes to detect vegetables, \ac{MOTS} provides instance segmentation image mask, which can be used to infer unique characteristics and phenotype of individual vegetable, such as its size and shape, to determine optimal doze of chemical spray. 

In this paper, a novel method of \ac{MOTS} is proposed for instance segmentation and tracking of multiple vegetable plants for precision spray application of agricultural robots. The overview of the method is shown in \cref{fig:overview}. To tackle the challenging problem of associating vegetables with similar color and texture in consecutively captured images, their shape features, which consist of contour and blob features, are extracted to describe a unique feature for each individual vegetable and match the same vegetables in different images. By assigning a unique ID for each vegetable, it ensures the robot to spray each vegetable exactly once, while traversing along the farm rows. Compared to the conventional \ac{MOTS} methods, the proposed method is able to re-identify objects which have gone out of the camera field for a long time and re-appear again. With the proposed plant shape feature and data association strategy, the plant which re-occurs into the camera field of view is re-identified and its ID on its first occurrence is recovered. It is common for most of agricultural robots to travel backward when it needs to avoid unexpected obstacles or casually move out for refilling or recharging. Therefore, this is important to ensure each vegetable is sprayed only once, when the robot traverses back and forth and vegetables re-appear in the camera images.

The contributions of this paper are summarized as follows:
\begin{enumerate}
    \item [1)] Firstly, a novel \ac{MOTS} method is proposed for tracking and segmentation of vegetable plants for robotic precision spray. Based on the instance segmentation of each vegetable plant, shape information of each plant is extracted and matched for tracking individual plants. Specifically, the contour information of a plant represented by \ac{FD} and blob information of a plant represented by parameters of the fitted ellipse are used to uniquely identify the plant. Shape features of all tracks are stored, and during data association, not only the plants in the current active tracks are searched, but also plants which have gone out of the camera field of view but geographically close the plants in the current camera field of view are searched. As a result, the proposed method can effectively re-identify these re-occurred plants, and recover their previous IDs.
    \item [2)] A lettuce multi-object tracking and segmentation dataset, LettuceMOTS, is constructed and publicly released. It contains 12 sequences, 1308 RGB images with corresponding annotated labels, 314 object instances and 17562 masks. Based on the LettuceMOTS dataset, comprehensive experiments including ablation studies are conducted, which show the superior performance of the proposed method over two \ac{SOTA} \ac{MOTS} methods.
    \item [3)] The implementation of the proposed method is also publicly released for the benefit of the community.
\end{enumerate}

The rest of the paper is organised as follows.
In \cref{sec:related}, the related work of plant segmentation in agriculture and \ac{MOTS} methods are discussed.
In \cref{sec:method}, the details of the proposed method are illustrated.
In \cref{sec:dataset}, the details of data acquisition and structure of dataset are provided.
In \cref{sec:exp}, experimental validation of the proposed method, comparison against two \ac{SOTA} \ac{MOTS} methods, and ablation studies are presented.
\cref{sec:conclusion} presents conclusions and a discussion about further work.


\section{Related Work}
\label{sec:related}

Two main research fields related to the proposed method are accurate segmentation of plants and efficient tracking of them. Therefore, related work in terms of plant segmentation and \ac{MOTS} is presented in this section.

\subsection{Plant Segmentation}
\label{subsec:related:robotics_seg}

Plant segmentation requires the precise separation of plant from the background. 
Early work utilizes hand-crafted feature for crop segmentation~\citep{C_ISCID2015：Song}. However, hand-crafted feature needs to be designed and adjusted according to the specific application and situation, and is affected by factors such as illumination change.
In recent years, the emergence and application of \ac{DNN} have triggered fundamental changes in the field of computer vision. This is because the more advanced and representative features are extracted by a large number of convolutional layers and pooling operations. The perception ability of robots has also been greatly improved with the continuous advancement of \ac{DNN}. Recently, many methods based on deep learning have been proposed for plant segmentation and achieved impressive results~\citep{J_JFR2017:Bargoti, C_ICRA2018:Milioto}. 

\cite{J_JFR2017:Bargoti} deployed a ground vehicle to collect images in an apple orchard. Apple segmentation is conducted by utilizing multiscale multilayered perceptrons and \ac{CNN}, and the number of apples is counted with Watershed Segmentation and Cyclic Hough Transforms. The results show that the combination of Watershed Segmentation and \ac{CNN} achieves the best counting performance, and the square correlation coefficient is 0.826. 
~\cite{C_ICRA2018:Milioto} proposed a method for semantic segmentation of sugar beet utilizing vegetation indexes. The results show that it achieves image processing speed of 20Hz on a variety of robotic systems.
~\cite{J_ELEC2020:Khan} proposed CED-Net, a semantic segmentation approach to classify plant. This method is based on a cascaded encoder-decoder network, and outperforms other segmentation architectures at the time on four public agricultural datasets.
~\cite{J_COMPAG2021:Bai} deployed a multi-network model to solve the problem of cucumber segmentation and detection in multiple scenarios. They first utilized the improved U-Net~\citep{C_MICCAI2015:Ronneberger} method to perform pixel-level segmentation of cucumbers, and then performed the further detection with the object detection algorithm.

These methods can accurately segment plants and locate them with pixel-level accuracy. However, they do not solve the problem of tracking the same plant on consecutive frames. Traditionally, agricultural robots have to move a fixed distance, segment and spray all plants in the current camera field of view, and move to the next stop. It has to ensure either two adjacent camera field of views have minimum overlapping region and do not contain any same plant, so as to achieve the purpose of neither repeating nor missing any plant. However, it brings in extra harsh requirement of precise robot navigation, which is normally difficult to achieve for most robotic platforms in the challenging farm environment. 
This problem can be solved by \ac{MOT} or \ac{MOTS} technology, which assigns an ID to each plant for continuous tracking, and sprays each plant only once. With every plant being tracked, the precise navigation requirement is effectively released, and navigation becomes uncoupled with perception to most extent. 

\subsection{Multiple Object Tracking and Segmentation}
\label{subsec:related:mots}

The main goal of \ac{MOT} is to detect and associate the same object in an image sequence~\citep{J_AI2021:Luo}. Currently, these methods are divided into two categories, single-stage \ac{MOT} methods~\citep{ J_IJCV2021:Zhang, J_TIP2022:Liang} and two-stage \ac{MOT} methods~\citep{C_ICIP2016:Bewley, C_ICIP2017:Wojke, C_ECCV2022:Zhang}. 

SORT proposed by \cite{C_ICIP2016:Bewley} is a simple and fast tracking system. It predicts the position of targets in the current frame through the Kalman filter~\citep{J_1960KF}, and matches them with the Hungarian algorithm~\citep{J_2010Hungarian}. \cite{C_ICIP2017:Wojke} proposed DeepSort based on Sort, which integrates the appearance model to obtain the feature embedding of the object. It further solves the tracking failure problem caused by occlusion. The downside of the method is that it handles detection and feature extraction tasks separately, which slows down the its processing. \cite{C_ECCV2020:Wang} presented the first \ac{MOT} system that placed object detection and feature embedding in the same task network, and achieved near real-time running speed. By utilizing two homogeneous branches to perform detection and feature extraction tasks separately, \cite{J_IJCV2021:Zhang} overcame unfairness of the operation of the two tasks and achieved high detection and tracking accuracy. 

\ac{MOTS} extends the perception accuracy of \ac{MOT} further, by replacing the bounding box to pixel-wise instance segmentation. 
\cite{C_CVPR2019:Voigtlaender} first came up with the concept of \ac{MOTS} and proposed a baseline method named TrackR-CNN. It extends the Mask R-CNN~\citep{J_TPAMI2017:He} with three-dimensional convolution to combine contextual information and deploys association head to extract instance embedding for data association. 
\cite{C_ECCV2020:Xu}  proposed PointTrack, which performed the tracking-by-instance segmentation paradigm. It first obtains high-quality instance segmentation results with spatial embedding~\citep{C_CVPR2019:Neven}, and then extracts instance features from the segmentation results through an unordered 2D point cloud. 
Based on PointTrack, \cite{J_TIP2022:Gao} deployed SENet~\citep{J_TPAMI2017:Hu} as an instance segmentation network, and utilized IDNet to extract object embedding for lightweight and high efficiency. 
These methods show accurate and robust results in tracking cars and pedestrians of large variance in color and texture. However, to track plants in farms, which have similar color and texture, color and texture embedding is prone to failure. Furthermore, these methods discard objects which have gone out of the camera field of view after a long time, and assign new IDs to them if they re-occur again. For robotic precision spray application, this means repeated spray for the same plant. 

Specifically for agricultural application, several \ac{MOTS} methods have been successfully applied. ~\cite{J_RAL2022:Jong} presented a \ac{MOTS} dataset containing apple instances using wearable cameras and drone recordings. Experimental results of two open-source methods show that tracking apples with similar color and texture is challenging. 
\cite{J_agronomy2022:Qiang} proposed a tracking method for leafy plants. They first apply a weakly supervised instance segmentation of leafy vegetables through semi-supervised learning. Mask \ac{IoU} and bipartite graph matching are then used for data association and tracking. However, this method only uses the mask \ac{IoU} as a position feature, which is more similar to \ac{MOT}, and does not take full advantage of the information obtained by the pixel level instance mask.


\section{Method}
\label{sec:method}

\subsection{Feature Exaction}
\label{subsec:method:feature}

\begin{figure*}[ht]
  \begin{center}
  \includegraphics[width=\columnwidth]{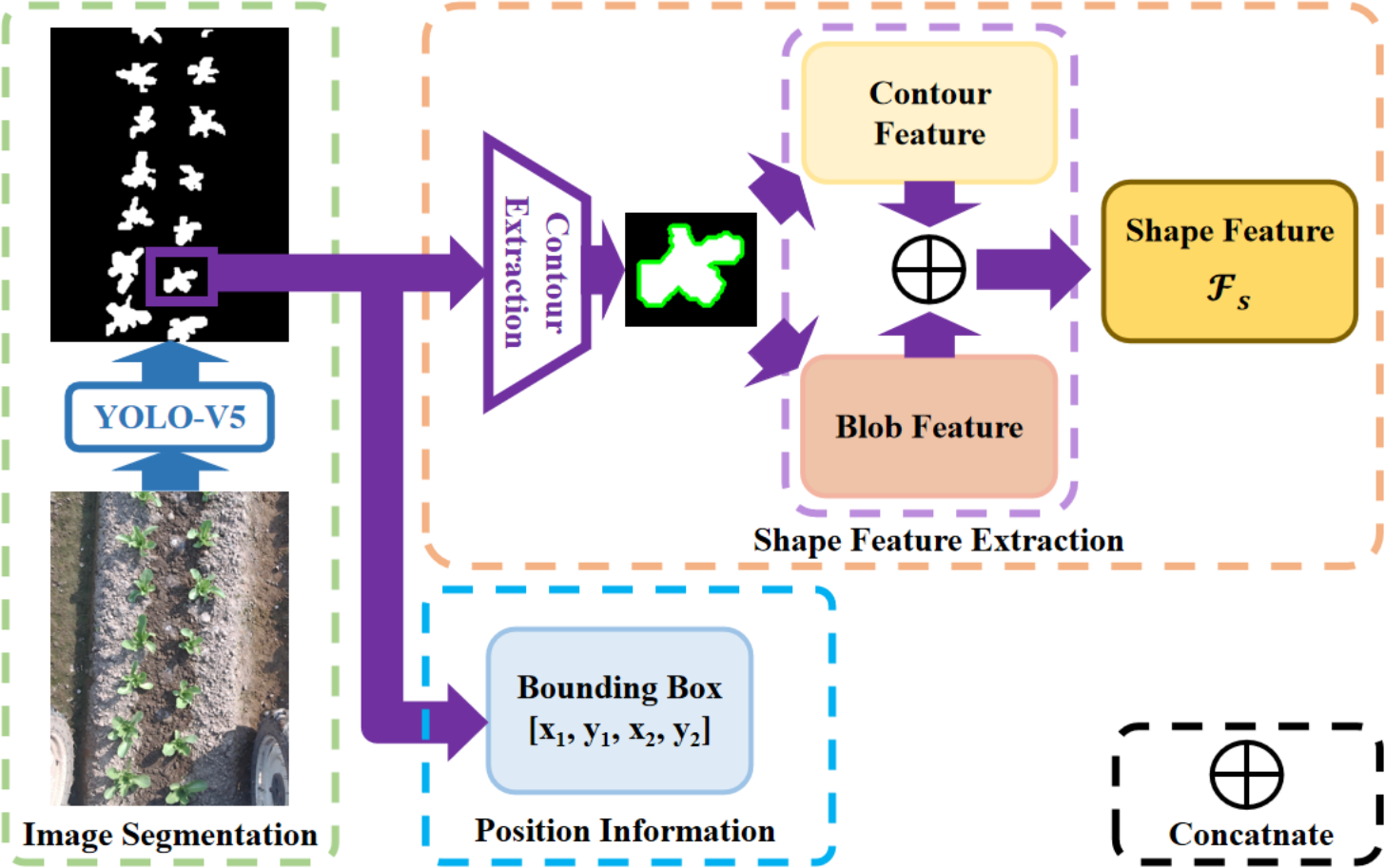}%
  \caption{Details of feature extraction. YOLOv5 instance segmentation yields masks and bounding box coordinates of all plants in the image. Based on the plant mask, the plant contour is extracted. Then \ac{FD} is applied to the plant contour to obtain contour feature, and an ellipse is fitted to the plant contour to obtain blob feature. The shape feature of the plant is obtained by combining the contour and blob features.
  }
  \label{fig:feature}
  \end{center}
\end{figure*}

The overview of the proposed method is shown in \cref{fig:feature}.
It adopts the famous YOLOv5 architecture to obtain instance segmentation of vegetables. YOLOv5 is preferred among various other instance segmentation nets due to its high accuracy and lightweight, which is important for the real time requirement for robotic operation.  The output size of the net is a matrix of $N \times H \times W $, where $N$ represents the number of objects, $H$ and $W$ are the height and width of the input image, respectively. The mask value of the plant is 1, and that of the background is 0. 

In agricultural scenarios, tracking of plants is generally difficult when conventional color and texture features are used \citep{J_FPS2022:Hu, J_RAL2022:Jong}. Our previous work \citep{J_FPS2022:Hu} presented a location information based feature extraction method based on the geometric relationship between the target plant and its neighboring plant. The method overcomes the challenging tracking problem of plants with similar appearance, but it requires presence of multiple plants on an image to extract such location feature. 
The proposed method pushes perception accuracy further to the pixel level, gaining more useful information such as plant shapes, which can be used to differentiate different plants.

The proposed method uses \ac{FD} to extract plant contour information based on instance segmentation mask of each plant. The \ac{FD} is a shape description method based on the Fourier transform of the shape contour and can represent the shape information in the frequency domain. In addition, blob feature of each plant is extracted by fitting an ellipse to its image mask. The blob feature is represented by the ratio of the major and minor axes, denoted as $R$, and center rotation angle, denoted as $\theta$. $\theta$ angle is normalized to keep $R$ and $\theta$ similar in magnitude. 
The combination of \ac{FD} of contour information and ellipse parameters of blob information serves as the shape information of the plant. It allows maximum information to be obtained with a small size number description dimension, which ensures less memory consumption. It is important for the shape information descriptor to be less in size, since descriptors of all plants are stored for tracking re-occurred plants.

As shown in \cref{fig:feature}. Firstly, contour of the plant is extracted based on its instance segmentation mask, using Suzuki85 border following algorithm \citep{J_CVGIP1985:Suzuki}. The contour descriptor \ac{FD}, and blob descriptor $R$ and $\theta$ are computed based on this. The extracted contour in $t$ frame is represented by its image coordinates as follows:

\begin{equation}
\label{eq_contour}
\mathcal{C}^t_n = \left\lbrace (x_{n}, y_{n})  |  n = 0, 1, \cdots, N-1\right\rbrace.
\end{equation}

The \ac{FD} derived in the form of centroid distance is used in this paper, which can better describe the shape features of the object~\citep{J_JVCIR2003:Zhang}. The centroid of the contour point is first calculated as follows,

\begin{equation}
\label{eq_cent}
x_c = \frac{1}{N} \sum_{n=0}^{N-1} x_{n},  y_c = \frac{1}{N} \sum_{n=0}^{N-1} y_{n}, \qquad n = 0,1,\ldots,N-1,
\end{equation}
where $x_c$ and $y_c$ are the X and Y coordinates of the contour centroid. The distance from the contour point to the centroid is computed as follows,

\begin{equation}
\label{eq_r}
r_n = \sqrt{(x_n - x_c)^2 + (y_n - y_c)^2}, \qquad n = 0,1,\ldots,N-1.
\end{equation}

Then, a discrete Fourier transform for the centroid distance $r_n$ is applied as follows,

\begin{equation}
\label{eq_DFT}
\Gamma_k = \frac{1}{N} \sum_{n=0}^{N-1} r_n e^{\frac{-j2\pi kn}{N}}, \qquad k = 0,1,\ldots,N-1.
\end{equation}

With a set of obtained $\Gamma_k$s composed of complex numbers, the contour descriptor \ac{FD} of the plant is computed by carrying out translation, rotation, and scale invariance operation as follows,

\begin{equation}
\label{eq_FD}
\bar{\Gamma}_i = \frac{\Vert \Gamma_k \Vert}{\Vert \Gamma_1 \Vert}, \qquad i=k-2, k = 2,3,\ldots,N-1.
\end{equation}
Note the first element $\Gamma_0$ is not used. Since the size of $\bar{\Gamma}_i$ is not fixed, the first $I$ elements of it is selected to represent its contour descriptor vector as follows, 
\begin{equation}
\label{eq_FD_I}
\hat{\Gamma}_I = 
\begin{bmatrix}
       \bar{\Gamma}_0 \\
       \vdots \\
       \bar{\Gamma}_{I-1}\\
     \end{bmatrix},
\end{equation}
In the baseline form of the proposed method, the first 5 elements, \textit{i.e.} $I=5$, are selected to make a balance between performance and speed. 

To extract the blob feature of the plant, an ellipse formulated below is fitted to plant mask,
\begin{equation}
\label{eq_ellipse}
A x^2 + B xy + C y^2 + D x + E y + F = 0,
\end{equation}
where $A$, $B$, $C$, $D$, $E$, and $F$ are the six parameters of the ellipse. When the $n$th contour point $(x_n, y_n)$ of the plant mask is considered, the corresponding fitting error $E_n$ is,

\begin{equation}
\label{eq_errfunc}
\begin{aligned}
E_n &= \left[\begin{matrix} x_n^2 & x_n y_n & y_n^2 & x_n & y_n & 1 \end{matrix}\right] \left[\begin{matrix} A \\ B \\ C \\ D \\ E \\ F \end{matrix}\right].
\end{aligned}
\end{equation}

Then the sum of squared errors of all contour points, $E_S$, is used to represent ellipse fitting error,
\begin{equation}
\label{eq_errsum}
E_S = \sum_{n=0}^{N-1} {E_n}^2, \qquad n = 0,1,\ldots,N-1.
\end{equation}
When, $E_S$ is minimized, we have the parameters of the optimum ellipse fitting.

The algebraic distance algorithm~\citep{C_BMVC1995:Fitzgibbon} is used to minimize the objective function $E_S$. Based on the obtained ellipse, the ratio $R$ of the long and short axes, and rotation angle $\theta$ of the ellipse are used to construct the blob feature. 

Finally, the shape feature of the plant is the combination of the contour feature $\hat{\Gamma}_I$ in \cref{eq_FD_I} and the blob feature represented by $R$ and $\theta$ of the ellipse. It is formulated as follows,

\begin{equation}
\label{eq_shapef}
\mathcal{F}_s = 
\begin{bmatrix}
       \hat{\Gamma}_I \\
       R\\
       \theta
     \end{bmatrix},
\end{equation}

In addition to the proposed shape information of the plant, the position information of the plant in the image frame is also utilized, as many \ac{MOTS} methods do. The position information is specifically represented by parameters of the coordinates of the bounding box containing the plant mask, which is denoted as follows,
\begin{equation}
\label{eq_position}
\textbf{B} = \left[\begin{matrix}B_1^x & B_1^y & B_2^x & B_2^y\end{matrix}\right],
\end{equation}
where $B_1^x$, $B_1^y$ are the horizontal and vertical coordinates of the upper left corner of the bounding box, and $B_2^x$, $B_2^y$ are the horizontal and vertical coordinates of the lower right corner of the bounding box. 

During tracking process, the bounding box of a plant in the current frame is predicted by Kalman filter first. When data association is successfully carried out, the bounding box is updated accordingly. 
The shape feature of the successfully tracked plant is updated to that in the current frame.
Note that since the plant at top or bottom of the image does not appear completely, they are discarded to maintain the performance of tracking process.

\subsection{Data Association}
\label{subsec:method:association}

\begin{figure*}[ht]
  \begin{center}
  \includegraphics[width=\columnwidth]{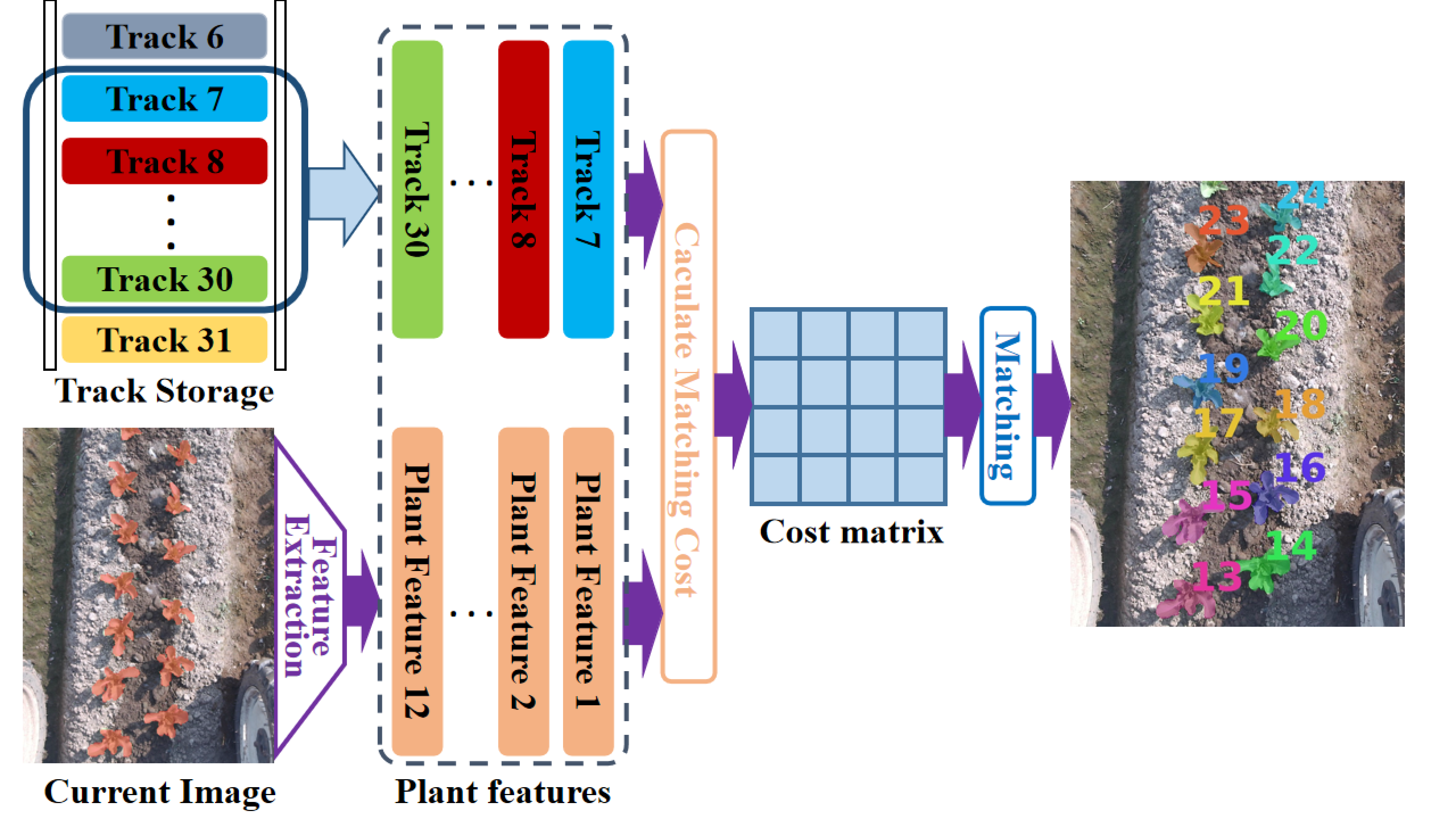}
  \caption{ Data association process. Shape features of plants in the current image frame are extracted. Meanwhile, tracks which are geographically close to the current active tracks, which are successfully matched to plants in the previous frame, are selected for match candidates. Matching cost is computed between plant features and tracks, and the matching cost matrix is constructed. Hungarian algorithm is used to resolve the optimum data association problem based on the matching cost matrix.}
  \label{fig:association}
  \end{center}
\end{figure*}

Data association is the process of matching the objects in two frames, and it is critical for tracking plants. It mainly includes two steps, which are calculating the matching cost between different objects and using the bipartite graph matching to associate objects according to the matching cost. 
The data association process is summarized in \cref{fig:association}. In the figure, the plant shape features are extracted by plant instances in the current image frame. Track refers to plants in the previously captured images frames, which has been successfully assigned unique IDs. After the data association process, a plant in the current frame either is assigned to a track if is successfully match to it, or initialized a new track if it is not matched to any previously constructed track.

In order to re-identify the re-occurred plant and recover its original ID, information of all tracks is stored in the proposed method, as opposed to only keeping active tracks in conventional \ac{MOTS} methods. However, objects of the current image frame are not matched against all tracks stored in the memory, but only matched to active tracks and their geographical neighbours, since there is no sudden jump for camera field of view. By effectively reducing the number of matching candidates, the execution time and chance of incorrect matching can be efficiently minimized. Since the IDs of tracks are initialized in numerical order while the robot traverses through the farm, geographically neighbouring plants have their IDs close to each other. There, the search scope of the tracks can be restricted within the range defined as follows,
\begin{equation}
\label{eq_search}
RNG = [ID_{min}-s, ID_{max}+s],
\end{equation}
where $ID_{min}$ and $ID_{max}$ are the minimum and maximum values of the successfully tracked object IDs in the previous frame, respectively. The variable $s$ controls the search scope and it is related to the maximum number of new objects that can potentially show up in the next frames. 
It is set to 6 in the implementation according to the speed of the robot, which ensures sufficient number of neighbouring plants are considered, and filters out plants that are too far away. 

To tackle the challenging problem of matching plants with similar appearance, establishing an effective matching cost is key to data association. 
The proposed matching cost consists of a position matching cost and a shape matching cost. 

Firstly, for position matching cost, \ac{GIoU} is adopted to calculate the position relationship between the two bounding boxes of plants as follows, 
\begin{equation}
\label{eq_giou}
GIoU = IoU - \frac{\left| {Area_C} - {Area_{Union}} \right|}{\left| Area_C \right|},
\end{equation}
where $Area_C$ is the area of the smallest rectangle containing two coordinate boxes, and $Area_{Union}$ is the area of the union of two bounding boxes. Since the range of \ac{IoU} is between 0 and 1, the position similarity based on \ac{IoU} is less distinguishable than \ac{GIoU} adopted by the proposed method. The value of \ac{GIoU} goes to -1 when the distance between two boxes is infinite, and goes to 1 otherwise. 
Then, the calculation of the position matching cost can be defined as follows,

\begin{equation}
\label{eq_poscost}
\delta_p = -G({B}^{t-1}_i, {B}^t_j),
\end{equation}
where $\delta_p$ is the matching cost of positions, $G$ represents the \ac{GIoU} in \cref{eq_giou} between the two bounding boxes. When calculating the position matching cost, Kalman filter is used to obtain the predicted position of the plant bounding box in the current frame. ${B}^{t-1}_i$ and ${B}^t_j$ represent the bounding box position predicted by Kalman filter of the plant in the $t-1$ frame, and the bounding box position of the detected plant in the $t$ frame, respectively. However, when a track is out of camera field of view, ${B}^{t-1}_i$ represents the position of the bounding box in the last frame before the plant disappears. 

For the shape matching cost, the cosine distance is utilized to indicate the cost score between two shape features, which is defined as follows,

\begin{equation}
\label{eq_shapcost}
\begin{aligned}
\delta_s & = D({\mathcal{F}^{t-1}_s}_i, {\mathcal{F}^t_s}_j) \\ & = 1 - \frac{{\mathcal{F}^{t-1}_s}_i  {\mathcal{F}^t_s}_j}{\Vert {\mathcal{F}^{t-1}_s}_i \Vert_2 \Vert{\mathcal{F}^t_s}_j\Vert_2} ,
\end{aligned}
\end{equation}
where $D$ refers to the cosine distance between the two plant shape features. ${\mathcal{F}_s}^{t-1}_i$ and ${\mathcal{F}^t_s}_j$ represent the shape feature vectors of two plants in $t-1$ frame and $t$ frame, as defined in \cref{eq_shapef}. 
Lower shape matching cost means two plants in the consecutive frames are more likely to be the same plant. When the two features are exactly the same, the cosine distance reaches 0. 

When the robot traverses on farm, the plant that disappears from the camera field of view is not tracked, and only plants detected in the current frame are tracked. For plants being actively tracked, both shape and position similarities are effective for matching, while only the shape similarity is effective for re-identifying the re-occurred plants. The overall matching cost combining both shape and position matching cost is formulated as follows,

\begin{equation}
\label{eq_cost}
    \delta = \delta_s (1 + \alpha \delta_p),
\end{equation}
where $\delta$ refers to the overall matching cost, and scalar $\alpha$ controls the influence of the position matching cost. For currently active tracks, the position information is more effective for matching, so $\alpha$ is set to be 1. For tracks that are currently inactive, their shape features are more effective and $\alpha$ is set to be 0.  

In addition, thresholds are applied to both position and overall matching costs to further filter out false positive matches whose position and overall matching costs are way too large. Threshold operation is formulated as follows,
\begin{equation}
\label{eq_matrix}
\delta =\left\{
	\begin{aligned}
	& \delta \quad & \delta < T_{all}\ and\ \ \delta_p < T_{p}\\
	& \infty \quad & otherwise
	\end{aligned}
	\right.,
\end{equation}
where $T_{all}$ and $T_{p}$ are maximum thresholds for overall and position matching costs, respectively. Based on empirical results, $T_{all}$ and $T_{p}$ in the implementation are set to 0.1 and 0.4, respectively. 

Finally, based on the overall matching cost $\delta$, Hungarian algorithm~\citep{J_2010Hungarian} is utilized for data association of multiple plants. 


\section{LettuceMOTS Dataset}
\label{sec:dataset}

In addition to the proposed \ac{MOTS} method, this paper presents and publicly releases a challenging lettuce \ac{MOTS} dataset, LettuceMOTS, captured by an agricultural robot. This section describes the agricultural robot and its sensor used in data acquisition and the structure of the dataset.

\subsection{Data Acquisition}
\label{subsec:dataset:acquisition}

\begin{figure}[htbp]
\centering
\subfloat[]{
    \label{fig:farm}
    \centering
    \includegraphics[width=0.45\textwidth]{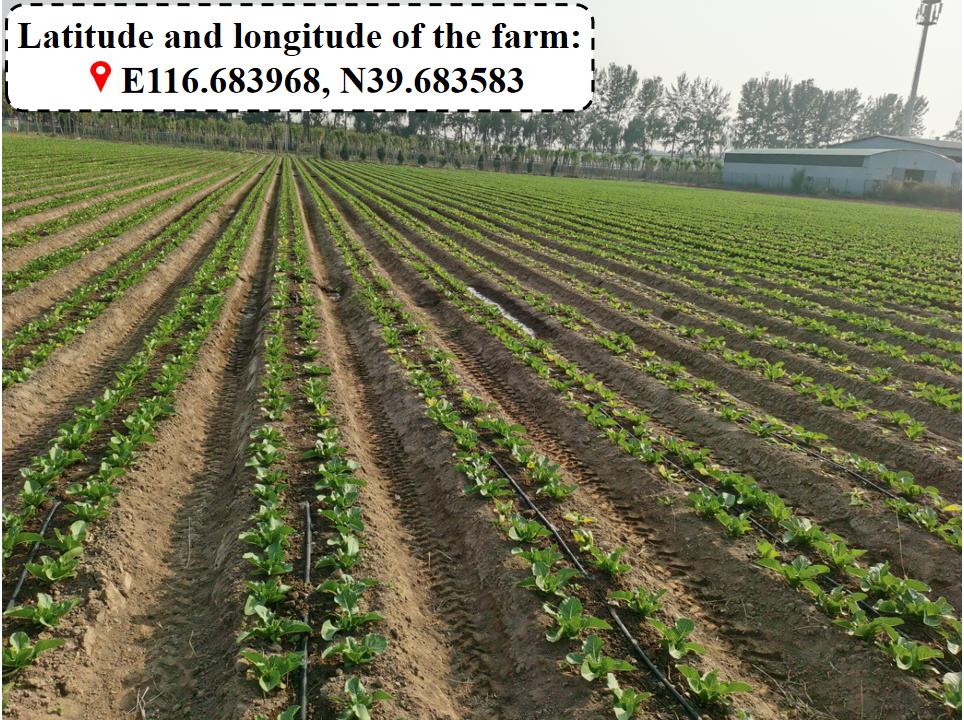}
}
\hspace{0.1in}
\subfloat[]{
    \centering
    \label{fig:robotic}
    \includegraphics[width=0.45\textwidth]{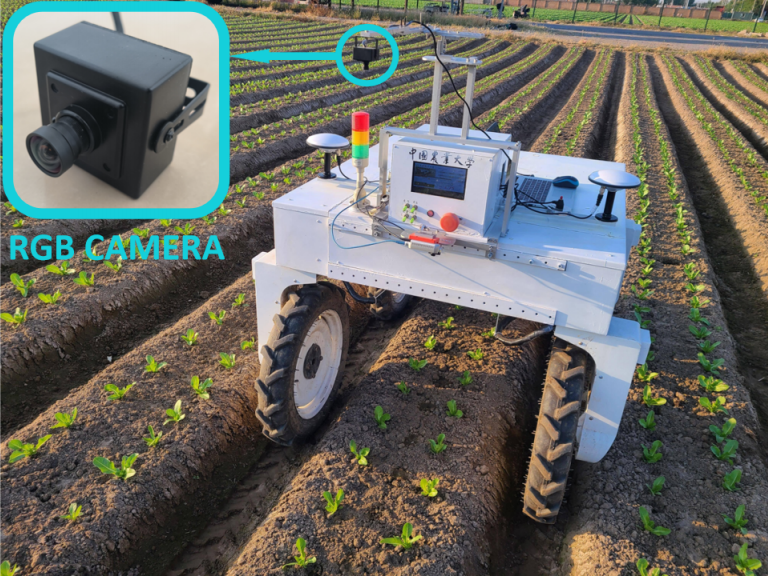}
}
\caption{Details of data acquisition. \cref{fig:farm} shows the lettuce farm where the data collection took place. \cref{fig:robotic} shows the setup of the data acquisition process based on VegeBot.}
\label{fig:acquisition}
\end{figure}

All images in LettuceMOTS were collected from a lettuce farm in Tongzhou District, Beijing, China, in September to October 2022 as shown in \cref{fig:farm}. 
The distance between two adjacent lettuces in the same row is about 0.15 m to 0.25 m, and the distance between adjacent rows is about 0.25 m to 0.3 m. The maximum weed density is close to 10 per square meter due to regular weeding.

Images are collected by VegeBot, a four-wheel-steer and four-wheel-drive agricultural robot designed and manufactured by China Agricultural University to perform autonomous operation in vegetable farms. The key parameters related to VegeBot are listed in \cref{tab_vegebot}. 
As is shown in \cref{fig:robotic}, the VegeBot is equipped with two vision sensors, and a RTK-GPS sensor for \ac{GNSS} based global localization. 
An Intel RealSense D435i depth camera with IMU sensor is mounted in front of the robot tilted downward for localization and autonomous navigation. A RGB monocular camera is mounted on top front location of the robot facing straight downward for segmenting and tracking vegetables for precision spray application.

The VegeBot can adopt two motion control modes: vision-based motion autonomous navigation and manual drive with a remote controller. In order to ensure the quality of the acquired images, the robot is remotely control during data acquisition process. 
Since the velocities and steering angle of the robotic four wheels can be controlled independently, the robot can move in a variety of motion modes. 
At the time of data collection, Ackerman steering is adopted for  the robot to travel straight forward or backward along the farm lanes. 

\begin{table*}[htp]
\caption{Key parameters of VegeBot}
 \begin{center}
 \label{tab_vegebot}  
 \centering
 \begin{threeparttable}
  \begin{tabular}{ c |  c }
   \hline
   Parameter  &  Value \\
   \hline
   Length  & 1.2 m\\ 
   
   Width    & 1.1 m \\ 
      
   Height  & 1.1 m \\ 
    
   Weight  & approx. 350 kg\\ 
   
   Max Load & 200 kg \\
   
   Max Speed  & 0.8 m/s\\ 
   \hline
  \end{tabular}
\end{threeparttable}
\end{center}
\end{table*}

The RGB camera used in this paper is installed approximately 1.4 m away from the ground. The camera has a 1/2.8-inch SONY IMX317 CMOS sensor. Its maximum resolution and pixel size are 3840 $ {\times}$ 2160 and 1.62 $\upmu$m ${\times}$ 1.62 $\upmu$m, respectively. The camera is able to capture images at 30 \ac{FPS} when the resolution is set to be 3840 $ {\times}$ 2160 or 120 \ac{FPS} when the resolution is set to be 1920 $ {\times}$ 1080.
The lens of the camera has a field of view (FOV) of 100 degrees and the f-number of 2.7.

\begin{figure}[htbp] 
   \centering 
   \subfloat[]{ 
   \centering
   \label{fig:921} 
     \includegraphics[width=0.3\textwidth]{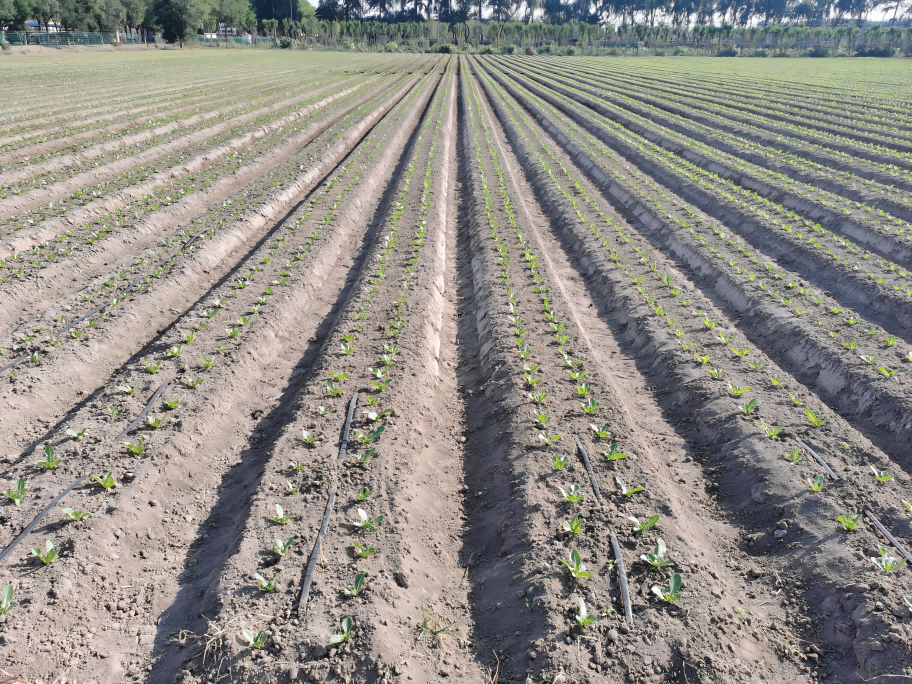} 
   }
   \subfloat[]{
    \centering
    \label{fig:929}
     \includegraphics[width=0.3\textwidth]{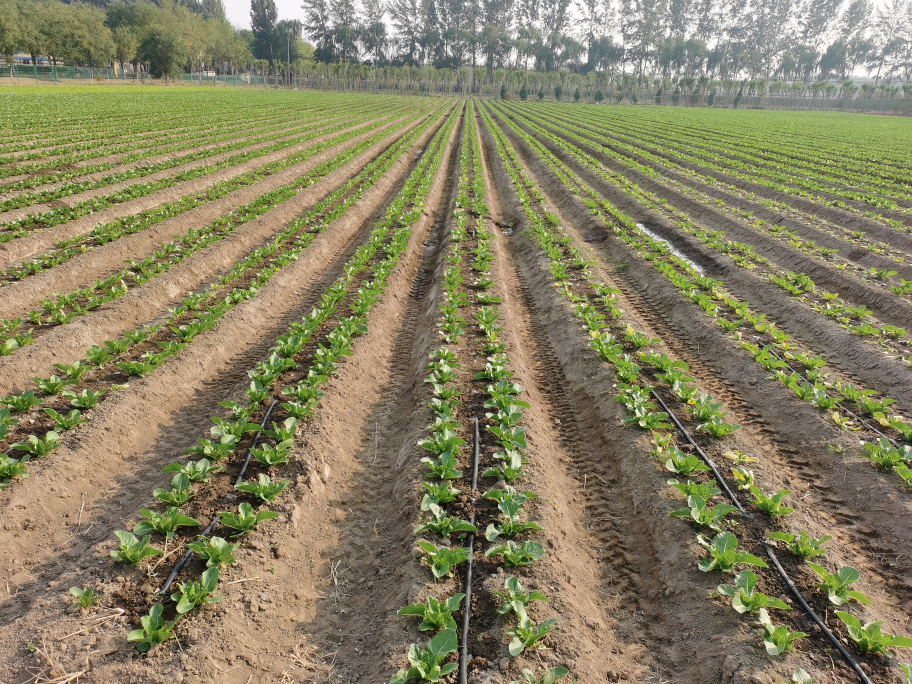}
   }
   \subfloat[]{
   \centering
   \label{fig:105}
     \includegraphics[width=0.3\textwidth]{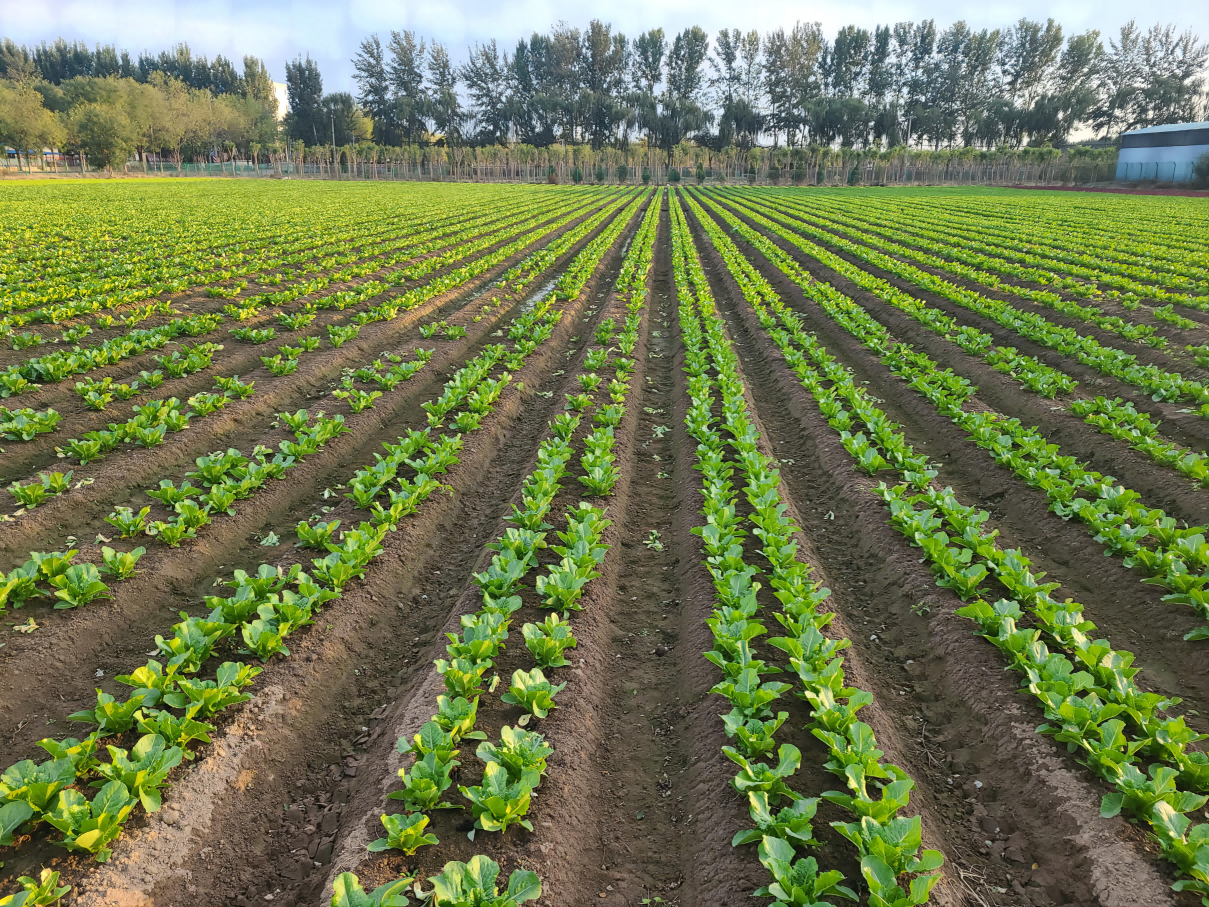}
   }\\ 
   \subfloat[]{
   \centering
   \label{fig:921img}
     \includegraphics[width=0.3\textwidth]{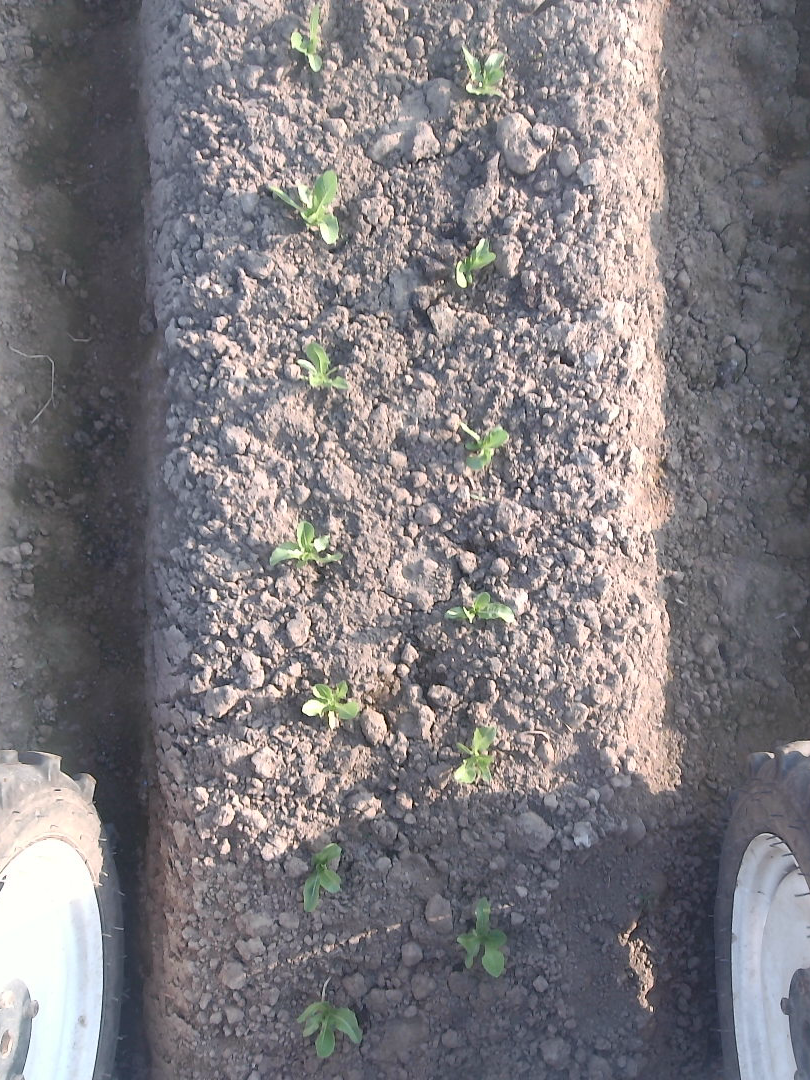}
   }
   \subfloat[]{
   \centering
   \label{fig:929img}
     \includegraphics[width=0.3\textwidth]{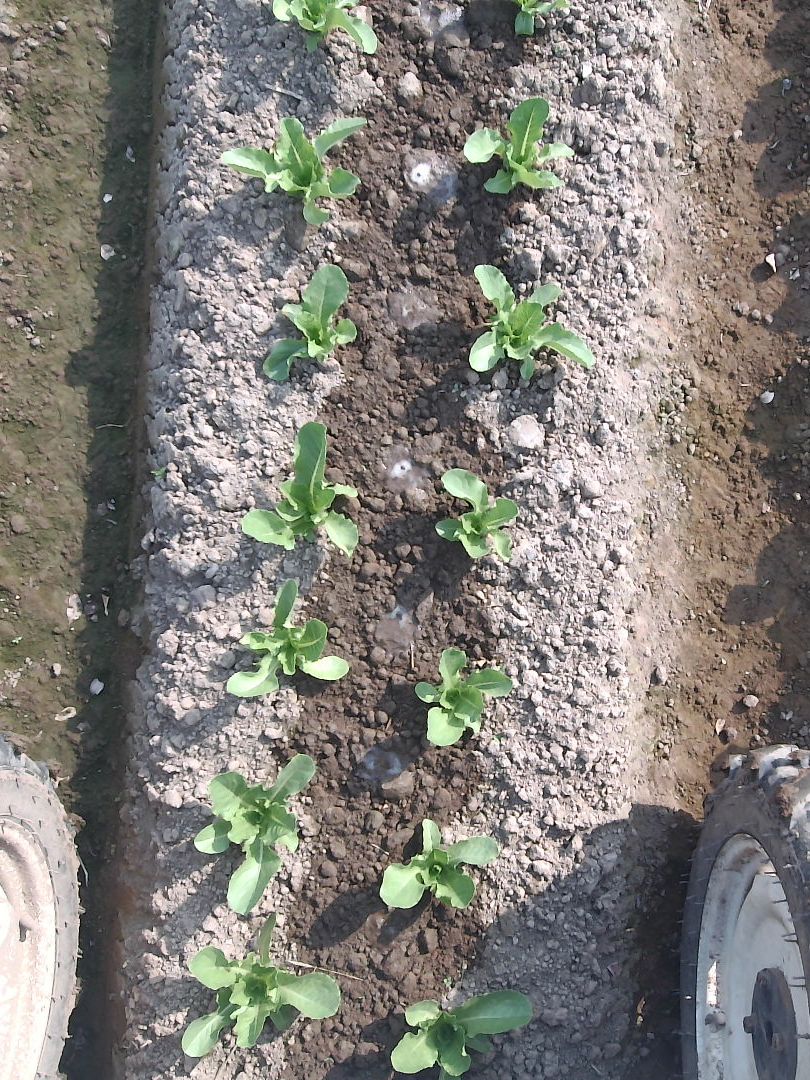}
   }
   \subfloat[]{
   \centering
   \label{fig:105img}
     \includegraphics[width=0.3\textwidth]{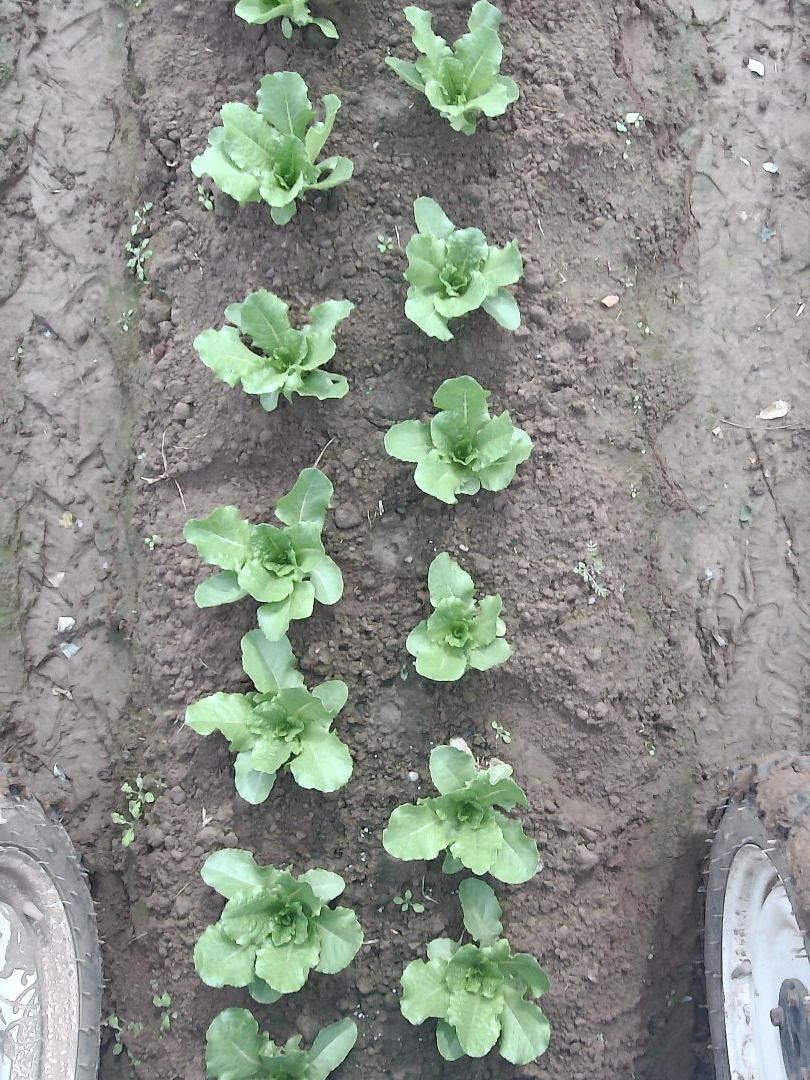}
   }
   \caption{Overview of the lettuce farm and typical captured images at 3 different growth stages of lettuces. Lettuces are at the growth stage of seeding, rosette, cupping and head in subfigures (a) and (d), (b) and (e), and (c) and (f), respectively.}
   \label{fig:growth img}
\end{figure}

Data collection is carried out at three different growth stages of the lettuce. Views of the lettuce farm and typical captured images at three periods are shown in \cref{fig:growth img}.
The robot travels at a mostly constant velocity within a range between 0.35 m/s to 0.4 m/s. 
The robot moves both forward and backward, and there are vegetables, which re-occur after they have gone out of camera field of view for a long time, as a result. This makes our dataset more challenging than existing \ac{MOTS} datasets. The camera captures images with a resolution of 1920$ {\times}$1080 at 15 \ac{FPS}, and images were cropped into the resolution of 810$ {\times}$1080 to remove irrelevant areas. Images are collected in natural light, and the exposure time of the camera is automatically determined. 

\subsection{Dataset Structure}
\label{subsec:dataset:structure}
The format and structure of LettuceMOTS follow the famous KITTI MOTS format~\citep{C_CVPR2012:Geiger, J_IJRR2013:Geiger}. The annotation tool \textit{CVAT}(\url{https://github.com/opencv/cvat}) is used to label the captured images. \textit{CVAT} is developed and open sourced by Intel, and its tracking, interpolation and fine tuning labeling method reduces manual labeling time significantly. 
The resulting annotation file of $TXT$ format has the following format,
\begin{equation}
	instance =\{frame_{id}, object_{id}, category, image_{height}, image_{width}, RLE\},
\end{equation}
where $frame_{id}$ is the ID number of the frame, $object_{id}$ is the ID number of the object, and $category_{id}$ is the ID number of the category of object. $category$ is 1 for vegetable.
$image_{height}$ and $image_{width}$ are the height and width of the image in which the plant is located. 
$RLE$ is a string of numeric encoding containing the plant mask information.

\begin{figure*}[!ht]
  \begin{center}
  \includegraphics[width=\columnwidth]{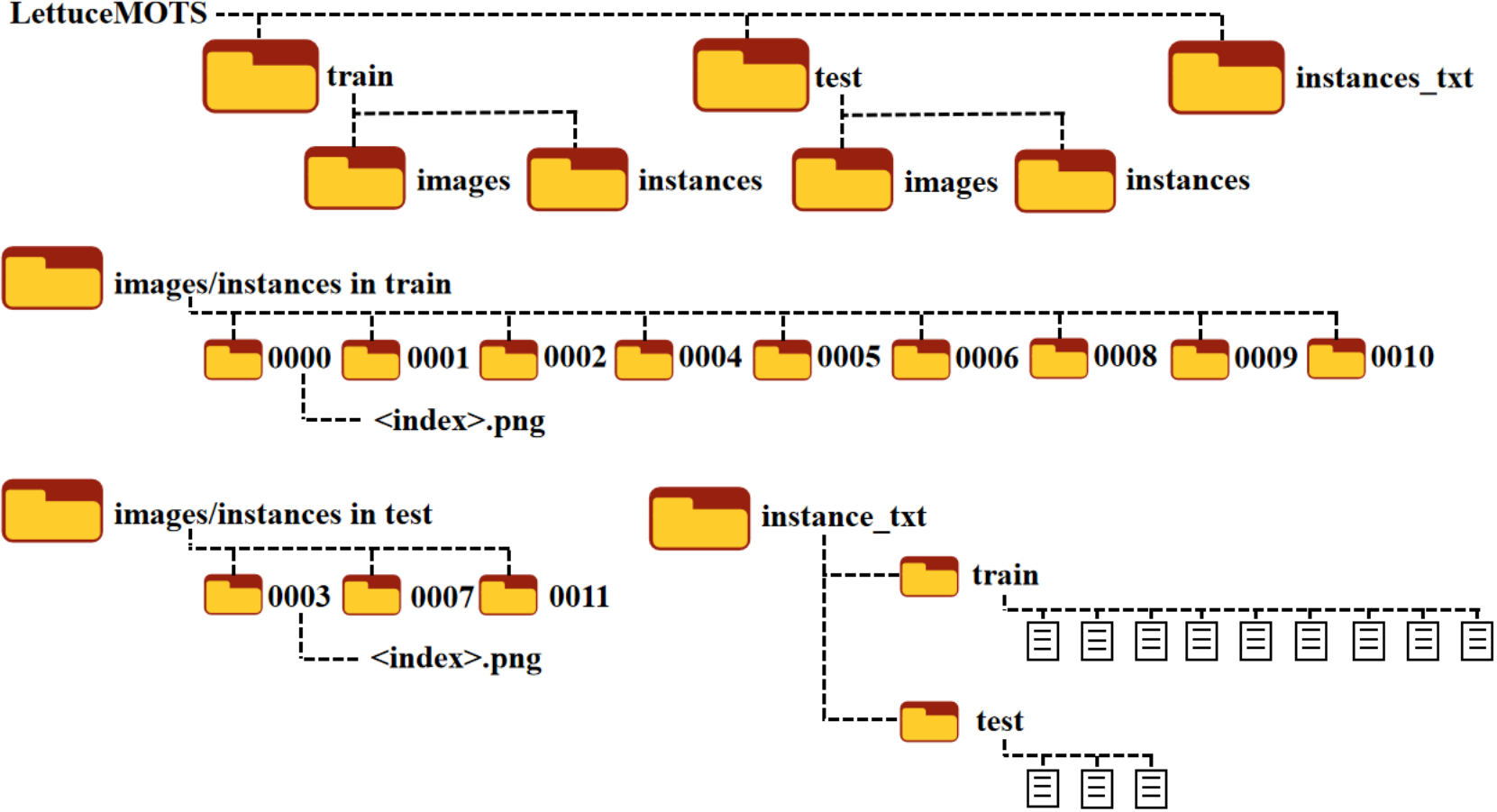}%
  \caption{Structure of LettuceMOTS dataset. The images and instances folders under train and test folders contain the captured RGB images and their corresponding instance segmentation masks for different image sequences. The instance\_txt folder contains \ac{MOTS} annotation for plant segmentation and tracking for different image sequences.}
  \label{fig:dataset}
  \end{center}
\end{figure*}

The file structure of LettuceMOTS is shown in \cref{fig:dataset}. It contains a total of 1308 frames, 314 objects, and 17562 manual annotated masks. The 12 sequences of the dataset are numbered from 0000 to 0011, with each growth period containing four sequences. Among them, three of them serve as the training set, and the other one serves as the test set. Detailed information about LettuceMOTS can be found in \cref{tab_dataset}.

\begin{sidewaystable}[htp]
\caption{Summary of the proposed LettuceMOTS dataset.}
 \begin{center}
 \label{tab_dataset}  
 \resizebox{\linewidth}{!}{
 \centering
 \begin{threeparttable}
  \begin{tabular}{ c | c  c  c  c | c  c  c  c | c  c  c  c | c}
   \hline
   Dataset sequences & $0000$ & $0001$ & $0002$ & $0003$ & $0004$ & $0005$ & $0006$ & $0007$ & $0008$ & $0009$ & $0010$ & $0011$ & Total\\
   \hline
   Length(Frame) &88  &50  &48  &245  &117  &51  &47  &244  &96  &54  &45  &223 &1308\\ 
   \hline    
   Plant Number &23  &23  &21  &40  &22  &22  &21   &37  &21  &23  &24  &37 &314\\ 
   \hline 
   Re-occurred Plants &9 &/  &/  &26  &10  &/  &/  &24 &8  &/  &/  &23 &/\\ 
   \hline
   Mask Number &1243 &678  &571  &3304  &1444  &669  &577  &3341 &1252  &733  &677  &3073 &17562\\ 
   \hline
   Robot Motion & F & F & F & B-F & F & F & F & B-F & F & F & F & B-F /\\
   \hline
   Growth Stage & \multicolumn{4}{c|}{Seedling stage} & \multicolumn{4}{c|}{Rosette stage} & \multicolumn{4}{c|}{cupping and head stage} & /\\
   \hline
   Weather & \multicolumn{4}{c|}{Sunny} & \multicolumn{4}{c|}{Sunny} & \multicolumn{4}{c|}{Cloudy} & /\\
   \hline
   Light intensity & \multicolumn{4}{c|}{Strong} & \multicolumn{4}{c|}{Strong} & \multicolumn{4}{c|}{Weak} & /\\
   \hline
   Date & \multicolumn{4}{c|}{Afternoon, Sept. 21, 2022} & \multicolumn{4}{c|}{Afternoon, Sept. 29, 2022} & \multicolumn{4}{c|}{Morning, Oct. 5, 2022} & /\\
   \hline
  \end{tabular}
  \begin{tablenotes}
    \item[1] {$F$ and $B-F$ in the Robot Motion refers to the robot motion of moving forward continuously, and the robot motion containing both forward and backward, respectively.}
  \end{tablenotes}
\end{threeparttable}}
\end{center}
\end{sidewaystable}


\section{Experimental and Results}
\label{sec:exp}

\subsection{Implementations Details}
\label{subsec:exp:details}

As mentioned in \cref{subsec:method:feature}, YOLOv5\citep{yolov5} is chosen to segment images. 
For segmentation of vegetables, the YOLOv5m model is selected to make a balance between accuracy and efficiency. It is fine tuned on the training set of the LettuceMOTS based on the pre-trained model based on COCO dataset provided by its author. SGD is used as the optimizer during training and the initial learning rate is set to be $1e^{-2}$. All other training parameters follow default parameters provided by YOLOv5.

For comparison purpose, two open source \ac{SOTA} \ac{MOTS} methods TrackR-CNN and PointTrack(The following open source implementations are used in the experiment. TrackR-CNN: \url{https://github.com/VisualComputingInstitute/TrackR-CNN} and PointTrack: \url{https://github.com/detectRecog/PointTrack}) are tested. They are fine tuned on the training set of LettuceMOTS based on the pre-trained model provided by the author. For PointTrack, learning rates of segmentation networks and tracker are set to $5e^{-6}$ and $2e^{-3}$, respectively. The learning rate of TrackR-CNN is set to be $5e^{-7}$. All other hyper parameters follow the default implementation.
All three methods, including the proposed method, are trained for 100 epochs to ensure fairness.
The training and inference of all methods are conducted on a computer with a NVIDIA GeForce RTX 2080Ti GPU and a Intel$^\circledR$ Core$^\text{TM}$ i7-10700K CPU.

\subsection{Evaluation Metrics}
\label{subsec:exp:evaluation}

The segmentation and tracking performance of the three methods are evaluated separately. The performance of instance segmentation is measured by \ac{AP}, which is widely used by many classic datasets, \textit{e.g.} COCO dataset~\citep{C_ECCV2014:Lin}. This metrics adopts $\ac{AP}$, $\ac{AP}_{50}$ and $\ac{AP}_{75}$ to show accuracy of segmentation. $\ac{AP}_{50}$ and $\ac{AP}_{75}$ are the $\ac{AP}$ at \ac{IoU} of 0.5 and 0.75, respectively. $\ac{AP}$ is the average of ten $\ac{AP}_{\ac{IoU}}$s, with \ac{IoU} ranging from 0.5 to 0.95 and an increase of 0.05 every step. 

The evaluation of \ac{MOTS} is relatively more complex than segmentation. In this paper, \ac{HOTA} proposed by \cite{J_IJCV2021:Luiten} is utilized for the evaluation of tracking tasks, which balances the performance of segmentation and tracking. \ac{HOTA} can better reflect the human's visual perception for \ac{MOTS} evaluation. It is calculated by \ac{DetA} and \ac{AssA} as follows,

\begin{equation}
\label{eq_hota}
HOTA = \sqrt{DetA \cdot AssA},
\end{equation}
where \ac{DetA} and \ac{AssA} represent comprehensive metrics of segmentation accuracy and association accuracy. The association metrics are defined as follows,

\begin{equation}
\label{eq_assa}
AssA = \dfrac{AssRe \cdot AssPr}{AssRe + AssPr - AssRe \cdot AssPr},
\end{equation}
where \ac{AssRe} reflects how good predicted trajectories cover ground truth trajectories, while \ac{AssPr} reflects the ability of predicted trajectories to continuously track the same ground truth trajectories. The detailed description of \ac{DetA}, \ac{AssA}, \ac{AssRe} and \ac{AssPr} can be found in the original work~\citep{J_IJCV2021:Luiten}, which is omitted here for the brevity of the paper. 

In this paper, we compute the above mentioned \ac{MOTS} evaluation metrics using the KITTI MOTS official kit(\url{https://github.com/JonathonLuiten/TrackEval}).

\subsection{Validation Results}
\label{subsec:exp:results}

\subsubsection{Results and Comparison}
\label{subsubsec:exp:results:comparison}

Firstly, we evaluate the segmentation results of the three \ac{MOTS} methods with \ac{AP} metrics, and results are shown in \cref{tab_seg}. It can be seen from the table that PointTrack has the highest $\ac{AP}$ score for segment accuracy. When \ac{IoU} is 0.5 and 0.75, YOLOv5 has the best segmentation performance among three methods. TrackR-CNN does not get the highest score, but achieves good and balanced result. 

\begin{table*}[htbp]
\caption{Segmentation performance of the proposed method and comparison to two state-of-the-art \ac{MOTS} methods.}
\begin{center}
\label{tab_seg}
 \resizebox{0.7\linewidth}{!}{
 \centering
 \begin{threeparttable}
  \begin{tabular}{c|c|c c c}
   \hline
  \multirow{2}{*}{Dataset} & \multirow{2}{*}{Method} &  \multirow{2}{*}{$AP$ $\uparrow$} & \multirow{2}{*}{$AP_{50}$ $\uparrow$} & \multirow{2}{*}{$AP_{75}$ $\uparrow$}\\ 
  & & \\ \hline
  \multirow{3}{*}{0003}  
  & TrackR-CNN & 0.592 & 0.967 & 0.770 \\
  & PointTrack  & \textbf{0.662} & 0.852 & \textbf{0.851}  \\
  & \textbf{Ours(YOLOv5)}  & 0.595 & \textbf{0.983} & 0.796 \\ \hline
  \multirow{3}{*}{0007}  
  & TrackR-CNN & 0.720 & 0.957 & 0.919 \\
  & PointTrack  & \textbf{0.824} & 0.940 & 0.940 \\
  & \textbf{Ours(YOLOv5)}  & 0.757 & \textbf{0.979} & \textbf{0.961}\\ \hline
   \multirow{3}{*}{0011} 
  & TrackR-CNN & 0.805 & 0.958 & 0.938 \\
  & PointTrack  & \textbf{0.852} & 0.968 & 0.954 \\
  & \textbf{Ours(YOLOv5)}  & 0.843 & \textbf{0.977} & \textbf{0.955} \\ \hline
  \end{tabular} 
  \begin{tablenotes}
    \item[1] Symbols $\uparrow$ after the evaluation metrics indicate the value of it is the higher the better. The bold numbers show the best performing method.
  \end{tablenotes}
\end{threeparttable}}
\end{center}
\end{table*}

Then, tracking performance of the proposed method is compared against the two \ac{SOTA} \ac{MOTS} methods with the test set of LettuceMOTS using the \ac{MOTS} metrics mentioned in \cref{subsec:exp:evaluation}. 
Results are shown in \cref{tab_mots}. It can be seen that our method yields superior performance than the other two methods in general. Specifically, our method gets the highest scores on all three test sets in terms of \ac{HOTA}, \ac{AssA} and \ac{AssPr}. It yields low \ac{DetA} scores, since the proposed method discards plants on top or bottom of the captured images that do not appear completely. As mentioned in \cref{subsec:method:feature}, we impose such constraint to minimize the false positive data association, especially for re-identifying re-occurred plants, since the plants with incomplete appearance show quite different shape feature.
PointTrack yields the lowest \ac{AssPr} because the same ID is assigned to multiple objects, and hence is not successful in tracking the same plant. Since PointTrack also introduces offset, position, and \ac{IoU} as clues during data association, ID switch rarely occurs and high \ac{AssRe} scores are achieved.
TrackR-CNN shows frequent ID switches when tracking the same object, and thus yields lower \ac{AssRe} scores.

\begin{table*}[htp]
\caption{Performance of the proposed method and comparison to two \ac{SOTA} \ac{MOTS} methods.}
\begin{center}
\label{tab_mots}
 \resizebox{\linewidth}{!}{
 \centering
 \begin{threeparttable}
  \begin{tabular}{c|c|c c c c c}
   \hline
  \multirow{2}{*}{Dataset} & \multirow{2}{*}{Method} &  \multirow{2}{*}{HOTA(\%) $\uparrow$} & \multirow{2}{*}{DetA(\%) $\uparrow$} & \multirow{2}{*}{AssA(\%) $\uparrow$} & \multirow{2}{*}{AssRe(\%) $\uparrow$} & \multirow{2}{*}{AssPr(\%) $\uparrow$}\\ 
  & & & & & \\ \hline
  \multirow{3}{*}{0003}  
  & TrackR-CNN & 50.016 & \textbf{77.757} & 32.332 & 57.965 & 39.485 \\
  & PointTrack  & 45.381 & 74.369 & 27.854 & 62.372 & 31.683 \\
  & \textbf{Ours}  & \textbf{71.989} & 70.827 & \textbf{73.561} & \textbf{74.792} & \textbf{84.318} \\ \hline
  \multirow{3}{*}{0007}  
  & TrackR-CNN & 59.918 & 84.195 & 42.709 & 55.134 & 63.978 \\
  & PointTrack  & 60.261 & \textbf{88.537} & 41.066 & \textbf{85.325} & 42.636 \\
  & \textbf{Ours}  & \textbf{72.083} & 71.419 & \textbf{72.843} & 73.565 & \textbf{89.998} \\ \hline
   \multirow{3}{*}{0011} 
  & TrackR-CNN & 62.042 & 88.480 & 43.543 & 61.389 & 63.052 \\
  & PointTrack  & 58.374 & \textbf{91.481} & 37.447 & \textbf{76.977} & 41.083 \\
  & \textbf{Ours}  & \textbf{70.095} & 68.868 & \textbf{71.433} & 71.597 & \textbf{95.167} \\ \hline
  \end{tabular} 
  \begin{tablenotes}
    \item[1] Symbols $\uparrow$ after the evaluation metrics indicate the value of it is the higher the better. The bold numbers show the best performing method.
  \end{tablenotes}
\end{threeparttable}}
\end{center}
\end{table*}

\begin{sidewaysfigure}[htbp]
  \begin{center}
  \includegraphics[width=\columnwidth]{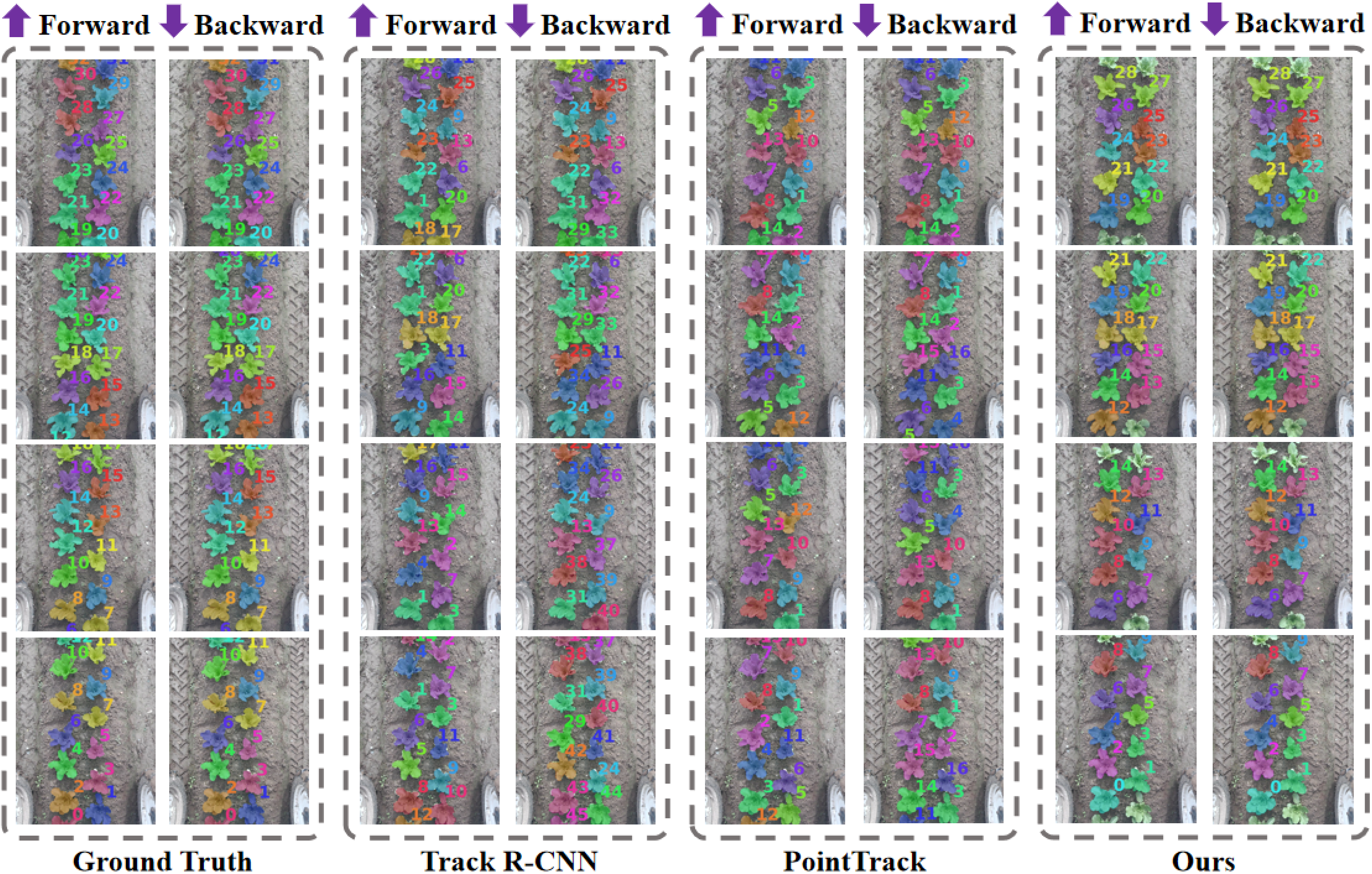}
  \caption{Tracking performance of three methods on test set 0011.}
  \label{fig:exp}
  \end{center}
\end{sidewaysfigure}

Qualitative examples of three \ac{MOTS} methods are shown in \cref{fig:exp}. 
We discuss the tracking ability of three methods with their performance on test set $0011$ as an example. 
The purple arrows above images represent forward and backward directions of the robot. The number in the upper right corner of each plant is assigned ID of that plant. 30 frames are skipped between every two neighbouring rows of images. For each method, the two images at left and right side are captured in the same positions when the robot travels forward and backward.
As can be seen from \cref{fig:exp}, TrackR-CNN and PointTrack yield false positive association of a plant which just goes out of camera field of view to a newly appeared plant. This is because these methods utilize instance embedding to associate objects. In our case, since vegetables are quite similar to each other in terms of color and texture, methods based on instance embedding are prone to false positive data association.
In addition, these two methods do not track plants which have gone out of camera field of view for a long time. As a result, they tend to assign new IDs when these plants appear again as the robot travels backward. It is reflected by plant IDs in the right columns of images of these two methods when the robot travels backward tend to be larger than those of the same plants in the left column of the images. 
We also can see that, PointTrack tends to mistakenly assign a previously assigned ID, which belongs to a plant previously appeared but has just gone out the camera field of view, to a newly appeared plant.  
Since PointTrack matches objects mostly based on color and texture and plants have quite similar color and texture, PointTrack mistakenly believes that the newly appeared plant is the plant which just has disappeared. 

In comparison, the proposed method yields superior data association performance, thanks to its extraction of vegetable shape feature which makes vegetable plants more differentiable with each other. In addition, limiting the searching range of tracks to those which are geographically close to current active tracks, \textit{i.e.} tracks associated to plants in the previous camera image, also contributes to reducing false positive matches and increasing data association accuracy significantly.

The inference speed of the three methods are shown in \cref{tab_time}. Our method yields the fastest running speed of approximately 29 \ac{FPS}. In terms of tracking speed alone, our method yields the highest processing speed exceeding 140 \ac{FPS}, since it is not a learning-based method. It can be run only on a CPU, which is easier to deploy on the robot.

\begin{table*}[!ht]
\caption{The inference speed of the proposed method and comparison to two \ac{SOTA} \ac{MOTS} methods.}
\begin{center}
\label{tab_time}
 \resizebox{\linewidth}{!}{
 \centering
 \begin{threeparttable}
  \begin{tabular}{c|c|c c c | c c c}
   \hline
  \multirow{2}{*}{Dataset} & \multirow{2}{*}{Method} &  \multicolumn{3}{c}{Time(ms) $\downarrow$} & \multicolumn{3}{c}{FPS $\uparrow$}  \\
  & & Segment & Tracking & Total & Segment & Tracking & Total
  \\ \hline
  \multirow{3}{*}{0003}  
  & TrackR-CNN & 581.67 & 20.49 & 602.16 & 1.72 & 48.83 & 1.66\\
  & PointTrack  & 69.39 & 44.90 & 114.29 & 14.41 & 22.27 & 8.75 \\
  & \textbf{Ours}  & \textbf{27.80} & \textbf{6.45} & \textbf{34.16} & \textbf{35.97} & \textbf{155.22} & \textbf{29.29} \\ \hline
  \multirow{3}{*}{0007}  
  & TrackR-CNN & 543.20 & 20.82 & 564.02 & 1.84 & 48.02 & 1.77\\
  & PointTrack  & 61.48 & 36.89 & 98.36 & 16.27 & 27.11 & 10.16\\
  & \textbf{Ours}  & \textbf{27.50} & \textbf{6.60} & \textbf{34.02} & \textbf{36.35} & \textbf{151.56} & \textbf{29.41} \\ \hline
   \multirow{3}{*}{0011} 
  & TrackR-CNN & 487.04 & 20.18 & 507.22 & 2.05 & 49.61 & 1.97\\
  & PointTrack  & 67.26 & 40.36 & 107.62 & 14.87 & 24.78 & 9.29 \\
  & \textbf{Ours}  & \textbf{28.07} & \textbf{7.00} & \textbf{34.93} & \textbf{35.64} & \textbf{143.05} & \textbf{28.62} \\ \hline
  \end{tabular} 
  \begin{tablenotes}
    \item[1] Symbols $\uparrow$ and $\downarrow$ after the evaluation metrics indicate the value of it is the higher the better or the lower the better, respectively. The bold numbers show the best performing method.
  \end{tablenotes}
\end{threeparttable}}
\end{center}
\end{table*}

\subsubsection{Ablation Studies}
\label{subsubsec:exp:results:ablation}

Finally, in order to validate the effectiveness of the proposed contour feature in terms of \ac{FD} of plant contour and blob feature in terms of $R$ and $\theta$ values of the fitted ellipse, we carry out an ablation study of the feature used. Specifically, performance of the proposed method using only contour feature and only blob feature is compared to using both of them, \textit{i.e.} the baseline of the proposed method. 

The results are shown in \cref{tab_ablation}. It can be seen that the baseline of the proposed method combining both contour and blob features yields the best performance compared to using only contour or blob feature in all three test sets of different growth stages. This validates that both contour and blob features are critical and effectively contribute to the performance gain brought by the proposed method.

\begin{table*}[htp]
\caption{Performance of the method using different plant shape features.}
\begin{center}
\label{tab_ablation}
 \resizebox{\linewidth}{!}{
 \centering
 \begin{threeparttable}
  \begin{tabular}{c|c|c c c c c}
   \hline
  \multirow{2}{*}{Dataset} & \multirow{2}{*}{Feature} &  \multirow{2}{*}{HOTA(\%) $\uparrow$} & \multirow{2}{*}{DetA(\%) $\uparrow$} & \multirow{2}{*}{AssA(\%) $\uparrow$} & \multirow{2}{*}{AssRe(\%) $\uparrow$} & \multirow{2}{*}{AssPr(\%) $\uparrow$}\\ 
  & & & & & \\ \hline
  \multirow{3}{*}{0003}  
  & Contour & 64.816 & \textbf{70.827} & 59.699 & 63.673 & 72.945 \\
  & Blob  & 65.325 & \textbf{70.827} & 60.668 & 63.956 & 75.993 \\
  & \textbf{Baseline}  & \textbf{71.989} & \textbf{70.827} & \textbf{73.561} & \textbf{74.792} & \textbf{84.318} \\ \hline
  \multirow{3}{*}{0007}  
  & Contour & 70.285 & \textbf{71.419} & 69.247 & 70.331 & 87.920 \\
  & Blob  & 64.633 & \textbf{71.419} & 58.579 & 62.295 & 77.420 \\
  & \textbf{Baseline}  & \textbf{72.083} & \textbf{71.419} & \textbf{72.843} & \textbf{73.565} & \textbf{89.998}  \\ \hline
   \multirow{3}{*}{0011} 
  & Contour & 66.485 & \textbf{68.868} & 64.278 & 66.303 & 87.968 \\
  & Blob  & 32.185 & \textbf{68.868} & 15.126 & 26.055 & 27.901 \\
  & \textbf{Baseline}  & \textbf{70.095} & \textbf{68.868} & \textbf{71.433} & \textbf{71.597} & \textbf{95.167} \\ \hline
  \end{tabular} 
  \begin{tablenotes}
    \item[1] Symbols $\uparrow$ after the evaluation metrics indicate the value of it is the higher the better. The bold numbers show the best performing method.
  \end{tablenotes}
\end{threeparttable}}
\end{center}
\end{table*}

Next, we study the influence of the length of \ac{FD} vector of the contour feature to the performance of tracking plants, by varying the length of \ac{FD} vector. Specifically, different lengths of \ac{FD} vector of 1, 3, 7, 9 are tested and compared with the baseline approach with the length of \ac{FD} vector of 5. 
Results are shown in \cref{tab_effect_shapef}, where $\ac{FD}_i$ indicates \ac{FD} with the length of $i$.
It can be seen that the performance of the method does not increase when the length of \ac{FD} vector is larger than 5. Therefore, the baseline configuration of the proposed method adopts \ac{FD} vector length of 5 to balance between accuracy and speed.

\begin{table*}[!ht]
\caption{Performance of the method with different lengths of \ac{FD} vector for plant contour feature.}
\begin{center}
\label{tab_effect_shapef}
 \resizebox{\linewidth}{!}{
 \centering
 \begin{threeparttable}
  \begin{tabular}{c|c|c c c c}
   \hline
  \multirow{2}{*}{Dataset} & \multirow{2}{*}{Feature} &  \multirow{2}{*}{HOTA(\%) $\uparrow$} & \multirow{2}{*}{AssA(\%) $\uparrow$} & \multirow{2}{*}{AssRe(\%) $\uparrow$} & \multirow{2}{*}{AssPr(\%) $\uparrow$}\\ 
  & & & &\\ \hline
  \multirow{5}{*}{0003}  
  & ${FD}_1$  & 56.982  & 45.970 & 50.302 & 66.101 \\
  & ${FD}_3$  & 69.360  & 68.186 & 70.079 & 80.894 \\
  & \textbf{Baseline (${FD}_5$)}  & \textbf{71.989} & \textbf{73.561} & \textbf{74.792} & \textbf{84.318} \\ 
  & ${FD}_7$  & \textbf{71.989} & \textbf{73.561} & \textbf{74.792} & \textbf{84.318} \\
  & ${FD}_9$  & \textbf{71.989} & \textbf{73.561} & \textbf{74.792} & \textbf{84.318} \\
  \hline
  \multirow{5}{*}{0007}  
  & ${FD}_1$  & 56.252  & 44.340 & 49.359 & 65.862 \\
  & ${FD}_3$  & 71.914  & 72.509 & 73.256 & \textbf{90.229} \\
  & \textbf{Baseline (${FD}_5$)}  & \textbf{72.083} & \textbf{72.843} & \textbf{73.565} & 89.998 \\ 
  & ${FD}_7$  & \textbf{72.083} & \textbf{72.843} & \textbf{73.565} & 89.998 \\
  & ${FD}_9$  & 71.656  & 71.983 & 72.873 & 89.210 \\
  \hline
   \multirow{5}{*}{0011} 
  & ${FD}_1$  & 50.500 & 37.125 & 44.436 & 61.218 \\
  & ${FD}_3$  & 69.960 & 71.157 & 71.322 & \textbf{95.167} \\
  & \textbf{Baseline (${FD}_5$)}  & \textbf{70.095} & \textbf{71.433} & \textbf{71.597} & \textbf{95.167} \\ 
  & ${FD}_7$  & \textbf{70.095} & \textbf{71.433} & \textbf{71.597} & \textbf{95.167} \\ 
  & ${FD}_9$  & \textbf{70.095} & \textbf{71.433} & \textbf{71.597} & \textbf{95.167} \\ 
  \hline
  \end{tabular} 
  \begin{tablenotes}
    \item[1] Symbols $\uparrow$ after the evaluation metrics indicate the value of it is the higher the better. The bold numbers show the best performing method.
    \item[2] $FD_i$ refers to taking the first $i$ element of \ac{FD} descriptors. For example, $FD_5$ takes the first 5 elements of \ac{FD} descriptors \textit{etc.}
  \end{tablenotes}
\end{threeparttable}}
\end{center}
\end{table*}

Another interesting question to be answered is whether we can achieve the same tracking performance by increasing the length of \ac{FD} vector of the contour feature and using such contour feature alone, \textit{i.e.} without using the blob feature. Therefore, we test tracking performance of the method using only contour feature of different lengths of \ac{FD} vector, and compare them with the baseline approach using the \ac{FD} length of 5 and blob feature. The results are shown in \cref{tab_effect_fd}, where $\ac{FD}_i$ w/o blob denotes the proposed method using the contour feature with \ac{FD} vector of length $i$ and without using blob feature. It can be seen from the results that the baseline approach of using both contour and blob feature still yields the best performance. Although increasing the length of \ac{FD} vector increases the tracking performance, the performance gain stops when length of vector reaches a certain number, which is not as good as the baseline approach in general.

\begin{table*}[htbp]
\caption{Comparison of tracking performance of different numbers of \ac{FD}s with baseline combined feature in the proposed method.}
\begin{center}
\label{tab_effect_fd}
 \resizebox{\linewidth}{!}{
 \centering
 \begin{threeparttable}
  \begin{tabular}{c|c|c c c c}
   \hline
  \multirow{2}{*}{Dataset} & \multirow{2}{*}{Feature} &  \multirow{2}{*}{HOTA(\%) $\uparrow$} & \multirow{2}{*}{AssA(\%) $\uparrow$} & \multirow{2}{*}{AssRe(\%) $\uparrow$} & \multirow{2}{*}{AssPr(\%) $\uparrow$}\\ 
  & & & & \\ \hline
  \multirow{6}{*}{0003}  
  & ${FD}_1$ w/o blob  & 33.389 & 15.818 & 24.912 & 26.390 \\
  & ${FD}_3$ w/o blob  & 52.820 & 48.413 & 52.409 & 69.232 \\
  & ${FD}_5$ w/o blob  & 64.816 & 59.699 & 63.673 & 72.495 \\ 
  & ${FD}_7$ w/o blob  & 66.699 & 63.272 & 66.515 & 76.267 \\
  & ${FD}_9$ w/o blob  & 66.699 & 63.272 & 66.515 & 76.267 \\
  & \textbf{Baseline (${FD}_5$ and blob)}  & \textbf{71.989} & \textbf{73.561} & \textbf{74.792} & \textbf{84.318} \\
  \hline
  \multirow{6}{*}{0007}  
  & ${FD}_1$ w/o blob & 26.712 & 10.025 & 18.818 & 19.118 \\
  & ${FD}_3$ w/o blob & 68.008 & 64.840 & 66.240 & 85.860 \\
  & ${FD}_5$ w/o blob & 70.285  & 69.247 & 70.331 & 87.920 \\ 
  & ${FD}_7$ w/o blob & 71.656  & 71.983 & 72.873 & 89.210 \\
  & ${FD}_9$ w/o blob & 71.656  & 71.983 & 72.873 & 89.210 \\
  & \textbf{Baseline (${FD}_5$ and blob)}  & \textbf{72.083} & \textbf{72.843} & \textbf{73.565} & \textbf{89.998} \\
  \hline
   \multirow{6}{*}{0011} 
  & ${FD}_1$ w/o blob & 21.064 &   6.462 & 17.438 &  9.513 \\
  & ${FD}_3$ w/o blob & 56.244 & 46.016 & 51.241 & 69.247 \\
  & ${FD}_5$ w/o blob & 66.485 & 64.278 & 66.303 & 87.968 \\ 
  & ${FD}_7$ w/o blob & \textbf{70.095} & \textbf{71.433} & \textbf{71.597} & \textbf{95.167} \\ 
  & ${FD}_9$ w/o blob & \textbf{70.095} & \textbf{71.433} & \textbf{71.597} & \textbf{95.167} \\ 
  & \textbf{Baseline (${FD}_5$ and blob)}  & \textbf{70.095} & \textbf{71.433} & \textbf{71.597} & \textbf{95.167} \\ 
  \hline
  \end{tabular} 
  \begin{tablenotes}
    \item[1] Symbols $\uparrow$ after the evaluation metrics indicate the value of it is the higher the better. The bold numbers show the best performing method.
    \item[2] $FD_i$ stands for taking the first few descriptors, for example, $FD_5$ is taking the first five descriptors as features \textit{etc.}
  \end{tablenotes}
\end{threeparttable}}
\end{center}
\end{table*}

\section{Conclusions}
\label{sec:conclusion}

To solve the challenging problem of associating vegetables with similar color and texture in consecutive images, in this paper, we propose a novel \ac{MOTS} method for segmenting and tracking of vegetables for robotic precision spray application in agriculture. The proposed method exploits shape feature of plants consisting of their contour and blob features, rather than conventional color and texture features, and yields superior tracking performance over conventional \ac{MOTS} methods in the challenging data association problem of vegetable plants with similar color and texture. In addition, the proposed method stores all constructed tracks, and searches within tracks which are all geographically close to vegetables in the current camera field of view during every data association step. Such a tracking strategy enables it to be able to re-identify re-occurred plants again, which is important to avoid spraying these plants more than once when the robot traverses back and forth. Comprehensive experiments and ablation studies are conducted to validate the superior performance of the proposed method, as well as various property of it. 
Furthermore, the dataset and implementation of the method are publicly released. Potential future work includes applying the proposed method to visual \ac{SLAM} and building an object level \ac{SLAM} system for robust localization and mapping in vegetable farms.



\end{document}